\documentclass[conference]{IEEEtran}
\IEEEoverridecommandlockouts
\usepackage{cite}
\usepackage{amsmath,amssymb,amsfonts}
\usepackage{algorithmic}
\usepackage{graphicx}
\usepackage{textcomp}
\usepackage{xcolor}
\usepackage{booktabs}
\usepackage{multirow}
\usepackage{tipa}
\usepackage{flafter}
\usepackage{url}
\def\BibTeX{{\rm B\kern-.05em{\sc i\kern-.025em b}\kern-.08em
    T\kern-.1667em\lower.7ex\hbox{E}\kern-.125emX}}

\begin{document}

\title{Comparing Human and Automatic Recognition of Dutch Dysarthric Continuous Speech: A Case Study\\
}

\author{
        \IEEEauthorblockN{
            Yuanyuan Zhang,
            Dimme de Groot, 
            Jorge Martinez,
            Odette Scharenborg 
        }
        \IEEEauthorblockA{\textit{Multimedia Computing Group} \\
        \textit{Delft University of Technology}\\
        Delft, the Netherlands \\
        \{y.zhang-44, d.c.c.j.degroot, j.a.martinezcastaneda, o.e.scharenborg\}@tudelft.nl
    }
}

\maketitle

\begin{abstract}
In collaboration with speakers with dysarthria, we are developing an application that supports speech communication of people with dysarthria by live captioning dysarthric speech and showing it as text on a device. To this end, our goal is to develop personalized dysarthric speech recognition (DSR) models. In this study, we compared the recognition performance of (1) naive human listeners, (2) three state-of-the-art, off-the-shelf automatic speech recognition (ASR) systems, and (3) two personalized DSR models on Dutch continuous read and spontaneous speech from a single speaker with severe dysarthria. Our results showed that both naive human listeners and the three off-the-shelf ASR systems exhibited word error rates (WERs) exceeding 70\% on average, indicating that DSR is highly challenging for both naive human listeners and ASR systems. The personalized DSR models, obtained by fine-tuning on the dysarthric speech, significantly reduced WER. Although overall WERs are still quite high (more than 23\%), the personalized DSR models outperformed the naive human listeners. These results show that personalized DSR can be useful for supporting day-to-day communication of speakers with dysarthria. 
\end{abstract}


\section{Introduction}
Communication is central to our participation in society and our well-being. For most of us, it comes naturally and is easy. However, this is not the case for millions of people worldwide who have dysarthria, a motor speech disorder caused by impaired muscular control of the speech mechanism, usually due to damage to the central or peripheral nervous system~\cite{darley1969differential}. It can result from a neurological disease or condition such as cerebral palsy~\cite{clement1959dysarthria}, Parkinson’s disease~\cite{kempler2002effect}, amyotrophic lateral sclerosis (ALS)~\cite{borasio1997palliative}, or due to stroke~\cite{mackenzie2011dysarthria} or traumatic brain injury~\cite{wang2005dysarthria}. In the United States, over one million people have Parkinson's disease~\cite{parkinson2026statistics}, with an estimated 44--88\% of them with dysarthria~\cite{asha_dysarthria}. In China, approximately eight million stroke patients are negatively impacted by dysarthria~\cite{liu2023audio}. 
An estimated 400,000 people in the Netherlands have dysarthria, based on the dysarthria prevalence numbers reported in~\cite{zoutenbier2016prevalentie} and the prevalence of the different causes of dysarthria. 

Speakers with dysarthria are among those who could benefit the most from speech technologies, such as automatic speech recognition (ASR): their everyday communication is often difficult or impaired due to their interlocutors having problems understanding their speech. This is both the case for naive listeners~\cite{bent2016individual,tjaden1995role, mengistu2011comparing, kim2016familiarization, sun2023cdsd, ogasawara2026ss} and for experienced listeners~\cite{tjaden1995role, kim2016familiarization, sun2023cdsd, green21_interspeech}. However, while ASR systems have improved significantly over the past decade and are now widely used in diverse applications~\cite{davitaia2025applications, wienrich2021trustworthiness, zhang2019life, luo2025assessing, elhadad2025improved, kumar2024comprehensive}, speakers with dysarthria have not benefited equally from this progress: even the best-performing commercial and publicly available ASR systems still achieved word error rates (WERs) of 49--60\% on English dysarthric speech~\cite{de2019impact, alsayegh2025zero} and WERs of 75--97\% on Dutch dysarthric speech~\cite{zhang2026}, showing that dysarthric speech remains challenging for state-of-the-art (SotA) ASR systems. 
Importantly, recent studies have shown that dysarthric speech recognition (DSR) performance can be substantially improved by training or fine-tuning ASR systems on dysarthric speech~\cite{yeo2022multilingual, liu2021recent, moro2019study, hasegawa2024community, martin2025project, nadrchal2026personalized, pokel2026adapt4me, huber2026adapting}.

In this paper, we investigate the potential of SotA DSR in supporting speech communication of speakers with dysarthria by comparing its performance to the recognition performance of naive human listeners for a single Dutch speaker with severe dysarthria. As such, we build on and extend previous research comparing human and automatic recognition of dysarthric speech~\cite{mengistu2011comparing, green21_interspeech, ogasawara2026ss, nadrchal2026personalized, sun2023cdsd}. 
The work is motivated by the strong belief that speakers with dysarthria should not be limited to communicating only with familiar interlocutors. They have both \textit{the need} and \textit{the right} to communicate effectively with unfamiliar (or naive) listeners. 
The current paper is part of a larger project in which we, together with speakers with dysarthria, are developing an application that supports their speech communication by live captioning their speech, i.e., by automatically recognizing it and showing it as text on a device.

One of the challenges for DSR is that the acoustic characteristics of dysarthric speech vary substantially between individuals, influenced by differences in underlying etiologies, severity, and stage of disease progression~\cite{green21_interspeech, bent2016individual, fougeron2010developing, christensen12_interspeech}. For example, even among speakers with the same etiology (cerebral palsy), the same severity (very low intelligibility), and the same gender (male), the same DSR model performed differently across individuals, with a word recognition accuracy difference between the best (65\%) and worst (42\%) recognized speaker of 23\%~\cite{vinotha2024enhancing}. This motivates the use of personalized (or \textit{speaker-dependent}) DSR models, which have been shown to consistently outperform speaker-independent models~\cite{sun2023cdsd, Sanders2002, cf31cecfbcaf42a1abd05eb908c3deaf, klausbridging}, especially for speakers with severe dysarthria~\cite{green21_interspeech, tobin2022personalized, hawley2007speech}. Given the current limitations of SotA ASR systems in generalizing across the highly variable dysarthric speech~\cite{de2019impact, alsayegh2025zero, green21_interspeech}, developing personalized DSR systems aligns with the goal of developing inclusive speech recognition: developing ASR for everyone~\cite{scharenborg2021inclusive}. 

Prior research, primarily focused on English~\cite{bent2016individual, tjaden1995role, mengistu2011comparing, kim2016familiarization, green21_interspeech}, Mandarin~\cite{sun2023cdsd}, and Japanese~\cite{ogasawara2026ss}, and on read speech, investigated \textit{human} recognition performance on dysarthric speech and showed the difficulty that human listeners have with recognizing dysarthric read speech. Nevertheless, a comparison of human and personalized DSR model performance on dysarthric speech by Mengistu et al. found that naive human listeners outperformed personalized DSR models on English read isolated words~\cite{mengistu2011comparing}. 
More recently, in 2021, Green et al. showed, for the first time, that personalized DSR models outperformed three speech-language pathologists (SLPs) on English dysarthric read phrases~\cite{green21_interspeech}. However, these expert listeners still outperformed the off-the-shelf Google Cloud model~\cite{green21_interspeech} on the read speech. 
A similar finding was reported for Mandarin read sentences~\cite{sun2023cdsd}, where personalized DSR models outperformed experienced listeners for five out of seven speakers with dysarthria on read speech. More recently, single-speaker studies on Japanese read dysarthric speech and on isolated one- to three-word Czech dysarthric utterances showed that personalized DSR models outperformed naive human listeners~\cite{ogasawara2026ss, nadrchal2026personalized}, but not listeners familiar with the speaker, such as family and friends~\cite{ogasawara2026ss} or the best-performing of the speaker's assistants~\cite{nadrchal2026personalized}.

We compare human listeners' performance with that of three off-the-shelf SotA ASR systems (Whisper-large-V3~\cite{radford2023robust, openai_whisper_large_v3}, Google Chirp 3~\cite{zhang2023google, google_speech_models}, and the Omnilingual ASR model~\cite{omnilingual2025omnilingual}) and two personalized DSR systems, developed by fine-tuning Whisper-large-V3 and the Omnilingual ASR model on speech from the target speaker in our study. We use three speech types: generic read, customized read, and, for the first time, spontaneous speech. Through comparisons between the human listeners' and ASR performance, we identify potential directions for improving DSR performance. Our results identify particular areas for improvement of our personalized DSR models and make a step towards better understanding the communicative challenges faced by the speaker with dysarthria, and towards improving day-to-day communication for them through speech technology.

\section{Methodology}

\subsection{Participants}

Twenty 
native listeners of Dutch (9 females, age range: 19-38 years; and 11 males, age range: 19-40 years) were recruited from the social and work circles of two authors of the paper. Participants were offered a voucher as a token of appreciation for participating in this study; one participant declined it. The study was approved by the ethical committee of our university. No participants reported hearing problems. All participants had no experience with listening to dysarthric speech, i.e., naive listeners. 

\subsection{Stimuli}
The stimuli for the listening and ASR experiments were taken from the DysOne dataset for personalized DSR model development~\cite{zhang2026dysone}. 
The DysOne dataset is a bilingual (native Dutch and non-native English) and bimodal (speech and video) dataset containing recordings of a 35-year-old male 
Dutch speaker with severe flaccid dysarthria resulting from traumatic brain injury due to high-impact trauma. In this study, we used only the speech data and the Dutch portion of DysOne, consisting of 3.3 hours of read speech and 0.4 hours of spontaneous speech.

For Dutch, DysOne contains three speech types: 
\begin{itemize}
    \item \textbf{Generic read speech:} Read prompts of three utterance length ranges (with a range between $(1, 5]$, $(5, 10]$, and $(10, 15]$ words) were selected from the Corpus Gesproken Nederlands~\cite{oostdijk-2000-spoken}. 
    \item \textbf{Customized read speech:} These read prompts were prepared by the speaker with dysarthria. The content is related to statistics lectures and is in line with the speaker's needs and interests. The word count was not controlled during recording.  The word-count range of the utterances was $[3, 15]$ (Mean: 7.5 words per utterance).
    \item \textbf{Spontaneous speech:} It was elicited through open-ended questions and statistics-related presentations (again in line with the speaker's needs and interests). Utterance length was not controlled. After recording, the word-count range was $[3, 15]$ (Mean: 7.5 words per utterance).
\end{itemize}
The DysOne dataset is under development and will be made publicly available.

188 Dutch utterances (with 686 unique words) were selected as stimuli for our experiments. The stimuli consisted of 121 generic read speech utterances (mean word count: 8.1, SD: 4.0), 27 customized read speech utterances (mean word count: 7.8 , SD: 2.3), and 40 spontaneous speech utterances (mean word count: 7.6, SD: 3.5). The remaining data were used as training and validation sets to fine-tune the pre-trained ASR models (see Section~\ref{subsec:ASRs}). Details about the training, validation, and stimuli sets are given in Table~\ref{tab:split}. 

\begin{table}[t]
    \centering
    \caption{Number of utterances in the training, validation, and stimulus datasets. Cust is customized read speech; Read is all read speech (generic + customized); Spon is spontaneous speech. The two isolated-word utterances and one 16-word utterance were due to reading errors.
    }\label{tab:split}
    \resizebox{\columnwidth}{!}{
\begin{tabular}{lllllll|lll}
\toprule
 \multirow{2}{*}{\textbf{Subset}} & \multicolumn{5}{c}{\textbf{Generic read speech}} & \multirow{2}{*}{\textbf{Cust}} &\multirow{2}{*}{\textbf{Read}} & \multirow{2}{*}{\textbf{Spon}} & \multirow{2}{*}{\textbf{Total}} \\
\cline{2-6}
 & \textbf{Word} & \textbf{1--5} & \textbf{5--10} & \textbf{10-15} & \textbf{16} &  &  &  &  \\
\midrule

\textbf{Training}   & 2 & 204 & 718 & 143 & 1 & 95 & 1163 & 141 & 1304 \\
\hline
\textbf{Validation} & 0 & 27  & 84  & 20  & 0 & 14 & 145  & 20  & 165  \\
\hline
\textbf{Stimuli}    & 0 & 40  & 40  & 41  & 0 & 27 & 148  & 40  & 188  \\
\bottomrule
\end{tabular}
    }
\end{table}

\subsection{Procedure of the listening experiment}
All listening experiments were conducted in a sound-attenuated booth using the same headphones (Sennheiser HD 200 Pro) and laptop. The user interface for the listening experiments was developed using Streamlit~\cite{streamlit}. Prior to the experiment, listeners received instructions, signed an informed consent form, and filled in a questionnaire for meta information, including gender, age, and familiarity with dysarthric speech.

The 188 utterances were divided into four lists of 47 utterances each. Each list contained ten generic read utterances with (1,5] words, ten with (5,10] words, and ten or eleven with (10,15] words, six or seven customized read utterances, and ten spontaneous utterances. Each list was assigned randomly to a listener and was listened to by five different listeners. For each listener, the stimuli were presented in randomized order. Listeners were allowed to listen to each stimulus as many times as they wished, and the total number of replays per stimulus was recorded. Listeners were instructed to type what they heard and to leave words they could not recognize blank. They did not receive any feedback on their transcriptions, as this could lead to learning effects and thus influence the results due to the pop-out effect~\cite{davis2005lexical}. After completing the experiment, participants were allowed to compare their responses with the ground-truth transcriptions, and all did so. 

\subsection{Automatic speech recognizers}\label{subsec:ASRs}
Three off-the-shelf SotA ASR systems were evaluated on the 188 stimuli:
\begin{itemize}
    \item \textbf{Whisper-large-v3}~\cite{radford2023robust, openai_whisper_large_v3} was selected because it has been reported to achieve the best performance on dysarthric speech in comparison to eleven typical ASR models, including Google Chirp 2~\cite{zhang2023google}, Microsoft Azure~\cite{microsoft_speech_languages}, and Meta's multilingual massive speech model~\cite{pratap2024scaling}, in a paper from 2025~\cite{alsayegh2025zero}. In our experiments, the Whisper task parameter was set to ``transcribe'', the language was set to ``Dutch'', and the temperature parameter was set to 0.
    \item \textbf{Google Chirp 3} is the most recent general-purpose ASR model released by Google~\cite{google_speech_models, zhang2023google} and was included because its predecessor, Chirp 2, achieved the best performance on diverse Dutch speech among twelve recent ASR models~\cite{zhang2026}, while Google Chirp 3 achieved the lowest WER among eight SotA ASR models on Dutch semi-spontaneous speech~\cite{de2026semi}. We ran Google Chirp 3 using synchronous recognition through the API. 
    \item \textbf{The Omnilingual ASR model}~\cite{omnilingual2025omnilingual} from Meta is a recently-released large-scale multilingual model, trained on 4.3 million hours of multilingual speech, including very large-scale European speech datasets with among others Dutch~\cite{koluguri2025granary, pratap2024scaling}. Omnilingual ASR supports recognition of more than 1,600 languages and has led to further substantial ASR performance increases for typical speech~\cite{omnilingual2025omnilingual}. To our knowledge, it has not yet been tested on dysarthric speech. For that reason we included it. Specifically, we used the \texttt{omniASR-LLM-300M-v2} model~\cite{bezzam_omniasr_llm_300m_v2}. 
\end{itemize}

The personalized DSR models are obtained by finetuning the Whisper and Omnilingual ASR (OmniASR) models on the DysOne training data. For both models, a two-fold speed perturbation was applied with factors of 0.9 and 1.1. For Whisper fine-tuning, Low-Rank Adaptation (LoRA)~\cite{hu2022lora} was employed. During decoding, beam search was used with a beam size of 10. Except for Google Chirp 3 (since it is with API), all experiments were conducted on an NVIDIA A40 GPU on our university's cluster.

\subsection{Evaluation}
To ensure consistency between human transcriptions and those of the ASR systems, all transcriptions were converted to lowercase, punctuation was removed (except for the apostrophe), extra spaces were removed, digits or numbers were written out in full, and all non-linguistic symbols, filler words, and other transcriptions of non-lexical sounds were removed. Obvious typing errors and spelling mistakes were corrected when the intended word was fully unambiguous, e.g., ``nouja'' $\rightarrow$ ``nou ja'' (well); ``vijfen'' $\rightarrow$ ``vijven'' (five).

Speech recognition performance is measured in word error rate (WER), computed as
\begin{equation}
\mathrm{WER} = \frac{S + D + I}{N} \times 100\%,
\label{eq:wer}
\end{equation}
where $S$, $D$, and $I$ are the numbers of substituted, deleted, and inserted words in a transcription, by either a human listener or an ASR model, relative to the reference (i.e., ground-truth) transcription, and $N$ is the number of words in the reference. Because insertions are counted, the WER can exceed 100\%. A lower WER indicates better recognition performance. 

For the recognition performance comparisons of the human listeners and ASR models, we conducted statistical significance tests using a paired nonparametric bootstrap test with 10,000 utterance-based samples and 95\% confidence intervals (CIs)~\cite{1326009, ferrer2024good}. A difference is considered statistically significant if the CI excludes zero~\cite{ferrer2024good}. We derived two-sided p-values from bootstrap samples by generating the null-centered distribution, calculated as the proportion of samples where the absolute difference was at least as large as the observed absolute difference~\cite{BoosStefanski}. 

To explore the effect of speech type and utterance length (word count) on the recognition performance of the human listeners and ASR models, a multi-factor linear model was applied using the estimated marginal means package (emmeans) in R~\cite{lenth2023emmeans}: 
\begin{equation}
\mathrm{WER}_{\mathrm{utt}} \sim \mathrm{system} \times \mathrm{type} \times \mathrm{wordcount}_{\mathrm{utt}},
\label{eq:lm}
\end{equation}
where $\mathrm{WER}_{\mathrm{utt}}$ is the WER of an utterance, $\mathrm{system}$ denotes the human listeners or an ASR model, $\mathrm{type}$ the speech type, and $\mathrm{wordcount}_{\mathrm{utt}}$ the number of words in the utterance.
Moreover, we conducted word-level and phoneme-level error analyses and investigated the effect of the number of replays of the stimuli on recognition performance.

\section{Results}
We first assessed whether the four experimental lists differed in recognition difficulty: A Kruskal–Wallis test showed no statistically significant differences between the four lists ($\chi^2(3)=2.29, p=.459$). 
Therefore, the results from the four lists were aggregated in the following analyses.

\subsection{Naive human listeners vs. ASR models}
Table~\ref{tab:model_comparison} presents the WERs of the human listeners averaged over all listeners, the off-the-shelf ASR models, and the fine-tuned ASR models on the three speech types. Additionally, the standard deviation (SD) for the human listeners is shown. 
First, both the human listeners and the off-the-shelf ASR models had substantial difficulty in recognizing the dysarthric speech, particularly for customized read speech, as shown by the high WERs. 
For the human listeners, individual WERs differed substantially, particularly for the customized read speech and spontaneous speech as is shown by the SDs in Table~\ref{tab:model_comparison}. The WER ranges of the best and worst performing human listener confirmed this large variability. For the generic read speech, customized read speech, and the spontaneous speech, the individual listener WERs ranged from 53.6\% to 80.4\% (difference of 26.8\%), 46.2\% to 114.1\% (67.9\% difference), 40.7\% to 90.0\% (49.3\% difference), respectively. These results indicate substantial variability in recognition performance across naive listeners. 


Comparison of the off-the-shelf ASR models showed that the Whisper model significantly outperformed the Google Chirp (95\% CI, [+8.197\%, +19.215\%], $p<.001$) and the OmniASR (95\% CI, [+11.099\%, +20.558\%], $p<.001$) models for generic read speech. No significant differences were found for customized read ($p>.104$) and spontaneous speech ($p>.097$). 
For all three speech types, the Whisper model performed similarly to the human listeners (generic read: 95\% CI, [-1.038\%, +8.817\%], $p=.115$; customized read: 95\% CI, [-41.414\%, -0.895\%], $p=.058$; spontaneous: 95\% CI, [-13.574\%, +3.133\%], $p=.265$). However, the human listeners significantly outperformed Google Chirp on generic read (95\% CI [+5.555\%, +19.864\%], $p=.0018$) and spontaneous speech (95\% CI [+5.275\%, +14.330\%], $p<.001$), and significantly outperformed the OmniASR model on all three speech types (generic read: 95\% CI [+8.047\%, +15.844\%], $p<.001$; customized read: 95\% CI [+9.347\%, +28.529\%], $p<.001$; spontaneous: 95\% CI [+2.289\%, +19.097\%], $p=.0144$).

\begin{table}[t]
    \centering
    \caption{WERs (\%) of the naive human listeners and off-the-shelf and fine-tuned ASR models.  
    The fine-tuned Whisper model and Omnilingual ASR model are denoted by FT-Whisper and FT-OmniASR.
    For the human listeners, the standard deviation of the WER is also given. Bold indicates the lowest WERs.}\label{tab:model_comparison}
    \resizebox{\columnwidth}{!}{
    \begin{tabular}{l|lll|l|l} 
        \toprule
        \multirow{2}{*}{\textbf{Model}} & \multicolumn{3}{c|}{\textbf{Read speech}} & \multirow{2}{*}{\textbf{Spon}} & \multirow{2}{*}{\textbf{Avg}}\\ 
         & \textbf{Generic} & \textbf{Customized}&\textbf{Avg} & & \\
        \midrule
        Humans & 69.2$\pm$7.8 & 90.1$\pm$13.7  & 72.9$\pm7.2$ & 70.1$\pm$12.0 & 72.3$\pm$7.3 \\
        \hline
        Whisper & 65.2  & 109.5 & 73.1 & 74.8 & 73.4 \\
        Google Chirp & 78.9  & 98.6 & 82.4 & 82.5 & 82.4 \\
        OmniASR & 81.0 &108.5&85.9  &80.4  & 84.8 \\
        \hline
        FT-Whisper & \textbf{26.1} & 26.5 & \textbf{26.2}  & \textbf{34.4} & \textbf{27.8} \\
        FT-OmniASR & 40.0 & \textbf{23.7}  & 37.1 & 42.4 & 38.2 \\
        \bottomrule
    \end{tabular}
    }
\end{table}

The best-performing model was the fine-tuned Whisper model (FT-Whisper). FT-Whisper significantly outperformed the fine-tuned OmniASR model (FT-OmniASR) on generic read speech (95\% CI, [+8.984\%, +18.919\%], $p<.001$), though no significant performance differences were found for customized read (95\% CI, [-12.617\%, +7.075\%, $p=.593$]) and spontaneous speech (95\% CI, [-0.332\%, +15.858\%], $p=.052$). 
Both models significantly outperformed the human listeners and the Whisper model for all speech types ($p<.001$). 

\subsection{The effect of speech type and utterance length}
As shown in Table~\ref{tab:model_comparison}, for human listeners, customized read speech was significantly more challenging than generic read speech ($t(546)=4.53, p<.001, d=0.966$) and spontaneous speech ($t(546)=2.63, p<.05, d=0.657$), which could be related to the topic of the customized speech, i.e., statistics-related content. Whisper showed significantly different performances for all three speech types (generic vs. customized: $t(546)=7.91, p<.001, d=-1.686$, generic vs. spontaneous: $t(546)=-3.318, p<.001, d=-0.608$, spontaneous vs. customized: $t(546)=4.316, p<.001, d=-1.078$). 
In contrast, FT-Whisper only showed a significant difference between generic read and spontaneous speech ($t(546)=-2.48, p=.036, d=-0.454$). Fine-tuning on the speaker’s speech thus eliminates the greater challenge of customized read speech for ASR systems. For both Whisper and FT-Whisper, the worse performance for spontaneous speech compared to read speech is consistent with previous findings for typical and diverse Dutch speech~\cite{feng2024towards, fuckner2023uncovering}. Note that human listeners performed similarly on generic read and spontaneous speech ($t(546)=-1.67, p=.212, d=-0.309$).

\begin{figure}[t]
    \centering 
        {\includegraphics[width=\columnwidth]{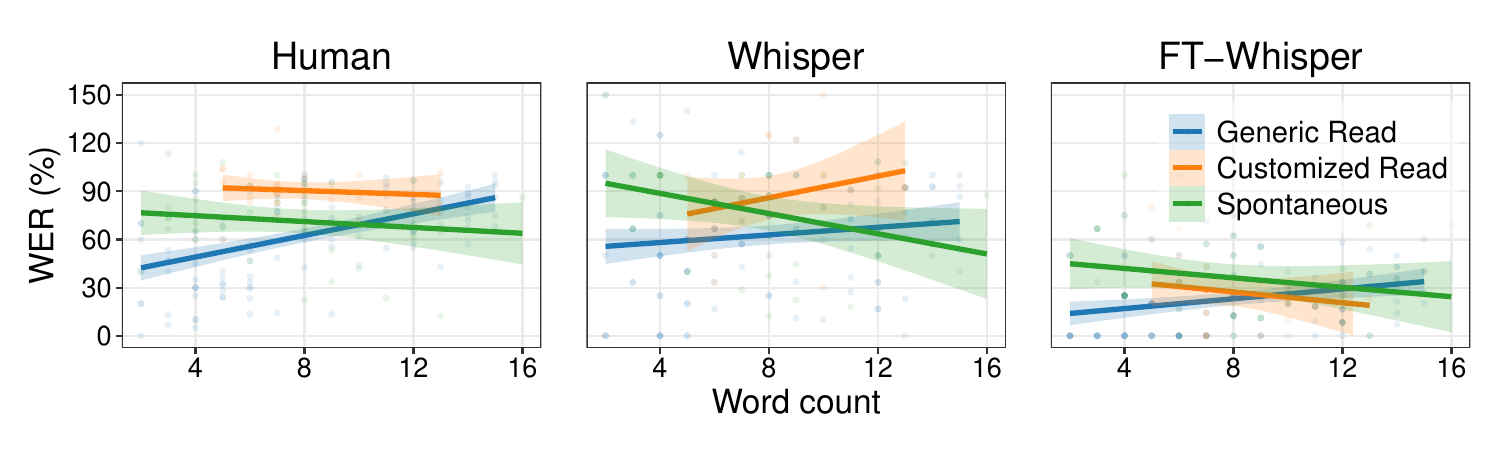}}
    \caption{Linear regressions of utterance length on WERs for naive human listeners, Whisper, and FT-Whisper models, for the three speech types separately.}\label{fig:uttlength}
\end{figure}

\begin{table}[t]
    \centering
    \caption{WERs (\%) of the naive human listeners and FT-Whisper on long generic read speech utterances (10–15 words, 41 utterances), and of Whisper on long spontaneous speech utterances (10–15 words, 9 utterances), calculated for the first 5 words of the utterance, the first 10 words of the utterance, and the full utterance. The table only shows results with a significant word-count effect in Fig.~\ref{fig:uttlength}.}
    \label{tab:long}
    
    \resizebox{\columnwidth}{!}{
    \begin{tabular}{llll}
        \toprule
        & \textbf{First 5 words} & \textbf{First 10 words} & \textbf{Full utterance} \\
        \midrule
        \multicolumn{4}{c}{Generic read speech} \\
        \hline
        Humans      & 77.2  & 78.3 & 78.4 \\
        FT-Whisper  & 26.3  & 28.5 & 29.4 \\
        \hline
        \multicolumn{4}{c}{Spontaneous speech} \\
        \hline
        Whisper &68.9 &68.9 & 67.3\\
        \bottomrule
    \end{tabular}
    }
\end{table}

Fig.~\ref{fig:uttlength} presents the effect of the number of words on recognition performance. We expected improved performance for longer utterances, as both humans and Whisper models can then, in principle, benefit from increased contextual information~\cite{radford2023robust}. However, this is not what we found. For customized read speech (orange lines), no statistically significant effect of word count was observed. 
For generic read speech (blue lines), we found a significant performance degradation with increasing number of words for the human listeners ($\beta=3.36$, $t(546)=4.98$, $p<.001$) and for FT-Whisper ($\beta=1.53$, $t(546)=2.26$, $p=.0242$). We hypothesized that these unexpected results may be attributed to fatigue of the speaker when producing long generic read utterances (10--15 words), resulting in less intelligible speech towards the end of long utterances. We tested this hypothesis by computing the WERs for the human listeners and the FT-Whisper model on the long generic read speech utterances with lengths of 10–15 words and by calculating the WER for the first five words, the first ten words, and the full utterance. The results are shown in the first block of Table~\ref{tab:long} and show that, for both human listeners and FT-Whisper, WERs increase slightly when more words of the utterance are included in the WER computation, indicating that later parts of long utterances contribute more to the overall WER. This finding is consistent with the speaker fatigue hypothesis. This explanation is further corroborated by the finding that for spontaneous speech (green lines in Fig.~\ref{fig:uttlength}) where the speaker has greater control over the utterance length of individual utterances, potentially reducing fatigue effects, performance improved with increasing word count, and significantly so for Whisper ($\beta = -4.70$, $t(546)=-3.50$, $p<.001$). As shown in the second block of Table~\ref{tab:long}, for the Whisper model, the later parts of long spontaneous utterances contribute less to the overall WER.

\subsection{Word and phoneme error analyses}

\begin{figure}[t]
    \centering
        {\includegraphics[width=1.0\columnwidth]{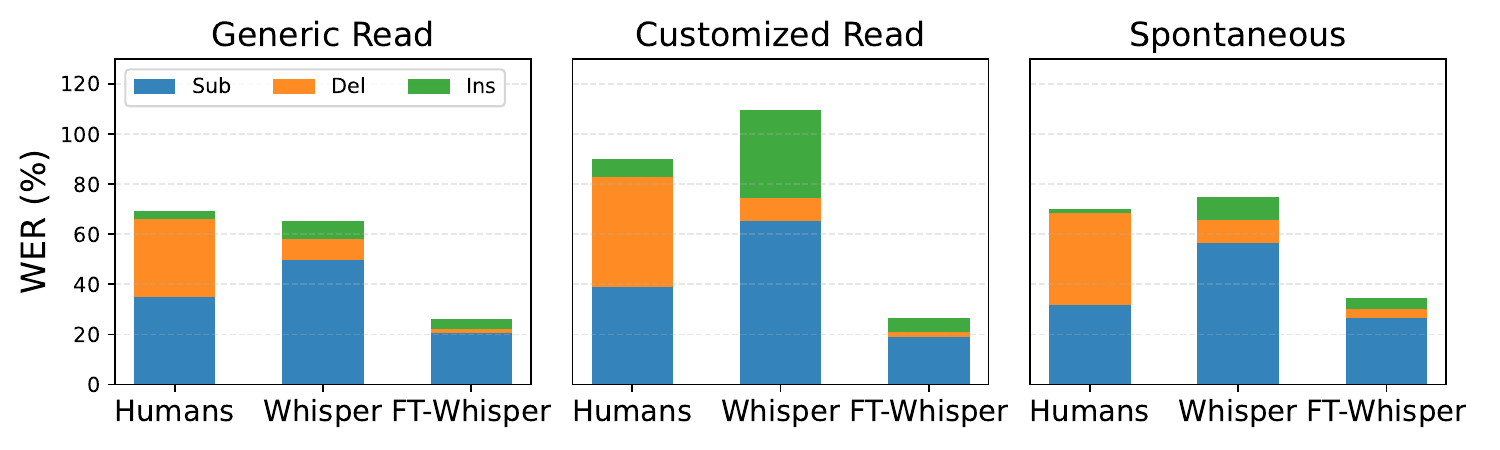}}
    \caption{Substitution, insertion, and deletion error rates of the naive human listeners, Whisper and FT-Whisper.}\label{fig:worderror}
\end{figure}

In the error analyses, we focus on the human listeners and the best off-the-shelf and fine-tuned models: Whisper and FT-Whisper. Fig.~\ref{fig:worderror} illustrates the substitution (blue bars), insertion (green bars), and deletion (orange bars) error rates for the human listeners, Whisper, and FT-Whisper for each of the three speech types. For human listeners, most errors were deletions 
and substitutions. 
For both Whisper and FT-Whisper, substitutions accounted for the majority of errors. In addition, the average transcription lengths of human listeners, Whisper, and FT-Whisper models are 5.5 words, 8.2 words, and 8.1 words, respectively; the average ground truth length is 7.9 words. 
This is in line with the instructions, which allowed listeners to leave out words they could not recognize, whereas ASR systems typically output something whenever speech is detected. Both the human listeners and FT-Whisper had relatively low insertion rates for all three speech types. In contrast, Whisper had a high insertion rate (35.1\%) for customized read speech when confronted with speech that deviates substantially from the typical speaker-based training material.

\begin{figure}[t]
    \centering
        {\includegraphics[width=\columnwidth, trim={0 4.5cm 0 1cm}, clip]
        {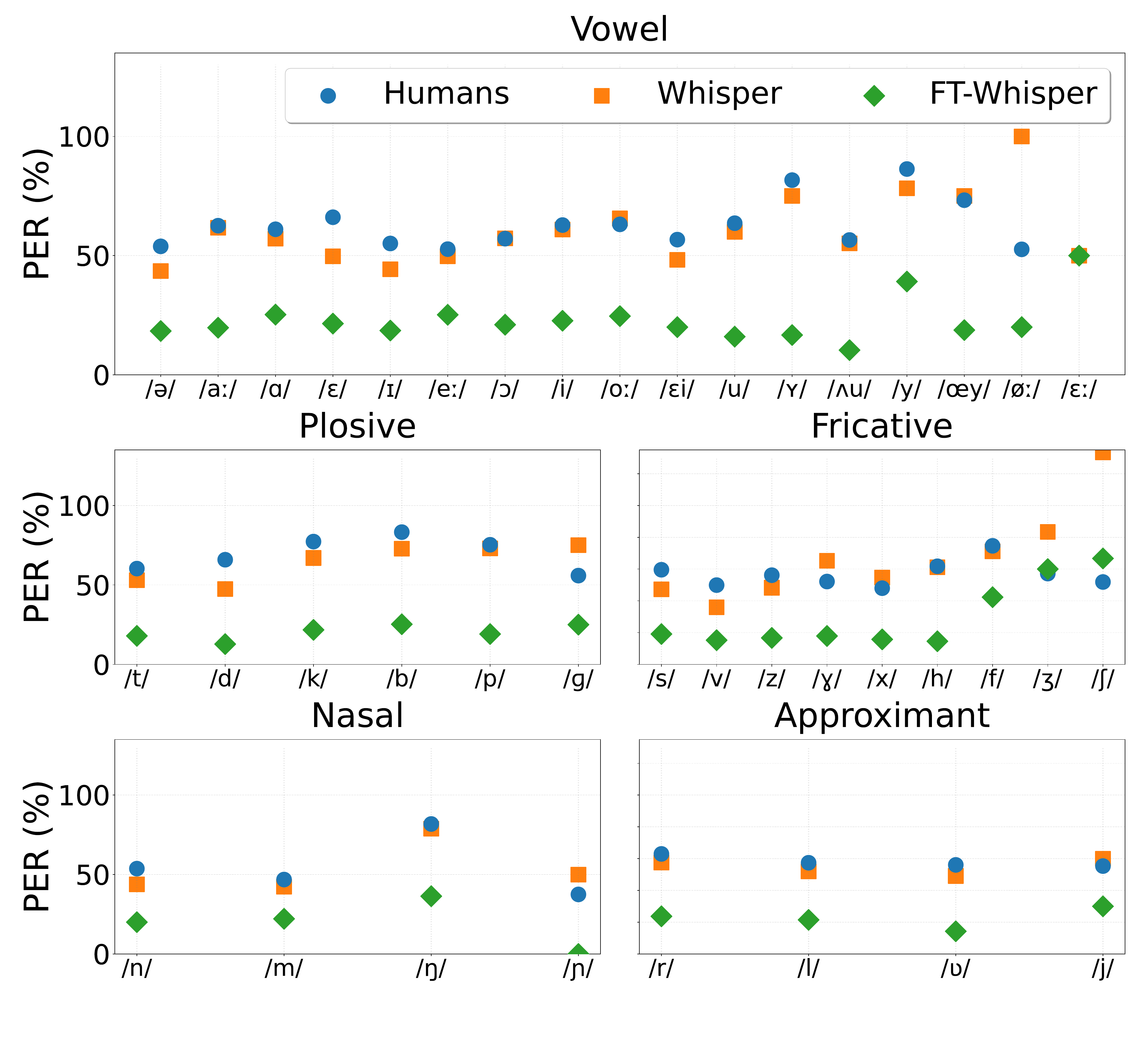}}
    \caption{PERs per phoneme, grouped by five articulatory manner classes. Phonemes on the x-axis are ordered by their frequency in the ground truth, with the left phoneme more frequent.}\label{fig:phoneme}
\end{figure}

Fig.~\ref{fig:phoneme} presents the results of the phoneme error analyses grouped by five classes of manner of articulation. 
Phoneme-level transcriptions were derived from the orthographic transcriptions using lexicon-based grapheme to phoneme mappings~\cite{halpern2022low, halpern_relative_phoneme_2020}. 
Overall, human listeners and Whisper show similar performance patterns for each of the five manner of articulation classes, while the phoneme error rates (PERs) of FT-Whisper are substantially lower. The largest differences are for individual phonemes. For particularly the $/$\textipa{S}$/$ and $/$\o\textlengthmark$/$, Whisper shows high PERs of 133.3\% and 100.0\%, respectively. Additionally, while fine-tuning substantially improves recognition performance for nearly all phonemes, it does less so for $/$\textepsilon\textlengthmark$/$. Despite fine-tuning, the error rate remains relatively high for two low-frequency phonemes ($/$\textipa{Z}$/$ and $/$\textipa{S}$/$) compared to those of human listeners. Additionally, $/$\textepsilon\textlengthmark$/$, $/$f$/$, $/$y$/$, $/$\textipa{N}$/$, and $/$j$/$ are the additional 5 phonemes with more than 30\% PER, indicating that, for this speaker with dysarthria, future ASR improvements could focus on these phonemes.

\subsection{The effect of listening times}

\begin{figure}[t]
    \centering
        {\includegraphics[width=0.7\columnwidth]{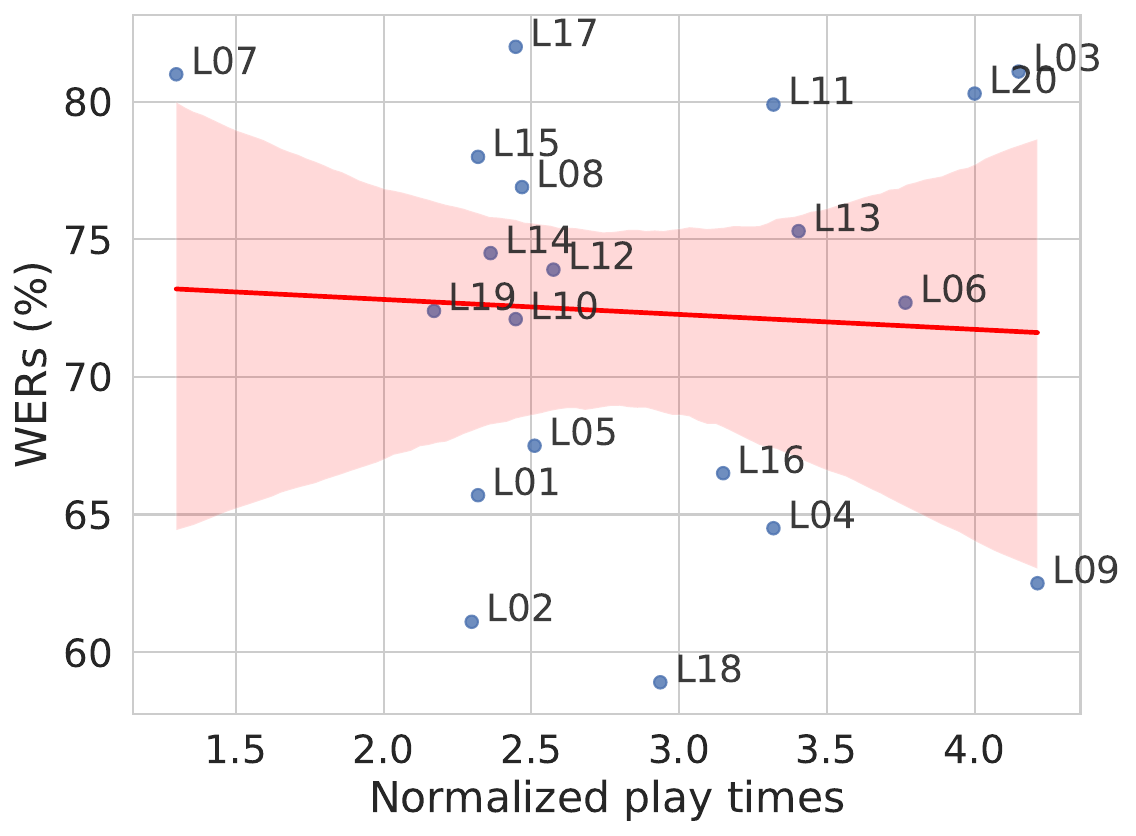}}
    \caption{Linear regression of normalized play times on WERs. $y = -0.54x + 73.90$ ($R^2 = 0.0033$). Shaded areas represent 95\% CIs.
    }\label{fig:LinearRegression}
\end{figure}
\begin{figure}[t]
    \centering
        {\includegraphics[width=\columnwidth]{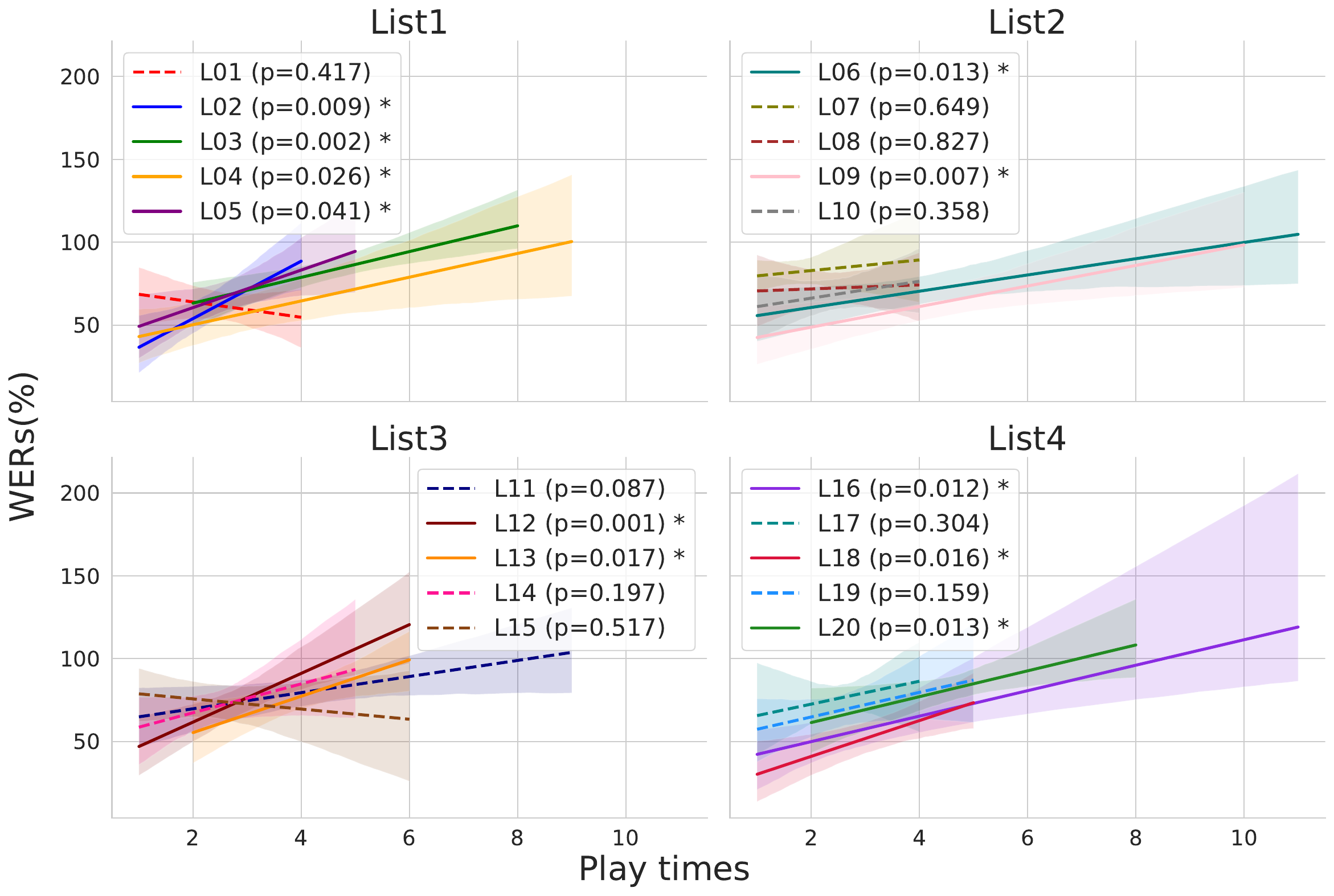}}
    \caption{Linear regressions of play times on WERs per utterance for twenty naive human listeners, separately. Shaded areas represent 95\% CIs. Statistically significant linear regressions ($p < .05$) are highlighted with solid lines and star marks in the legend, while non-significant ones are shown with dashed lines. 
    }\label{fig:LinearRegressionPerUtterance}
\end{figure}

Hilkhuysen et al.~\cite{hilkhuysen2012effects} found that recognition performance of human listeners on read speech from a single male speaker in background noise improved with increasing replaying times. We therefore analyzed whether in our study the number of times a participant listened to a stimulus had an effect on their recognition performance. As shown in Fig.~\ref{fig:LinearRegression}, 
a linear regression between the normalized number of play times per utterance for each human listener and the average WER for each listener, 
however, showed a non-significant effect ($p=.810$). We hypothesized that in our study, the number of play times may partly reflect the difficulty of recognizing dysarthric speech utterances. To verify this, for each human listener, a linear regression between the number of play times per utterance and the WER per utterance was plotted in Fig.~\ref{fig:LinearRegressionPerUtterance}. Results suggest that, for most listeners, the play times were higher for the more difficult utterances, as reflected by higher WERs, than for utterances with lower WERs confirming our hypothesis that more replays indicates more difficulty with the intelligibility of the speech.

\section{General discussion and conclusions}
In this paper, we compared the speech recognition performance of three SotA off-the-shelf ASR models, two newly trained personalized DSR systems, and naive human listeners on Dutch dysarthric continuous read and, for the first time, (a small set of) spontaneous speech from a single speaker with severe dysarthria. Specifically, we compared the recognition performance of Dutch native listeners with that of Whisper-large-V3, and the newly released Google Chirp 3 and Omnilingual models. 
Despite the SotA performance of Google Chirp 3~\cite{google_speech_models} and Omnilingual ASR~\cite{omnilingual2025omnilingual} on typical speech in diverse languages, they were significantly outperformed by the naive human listeners on Dutch dysarthric speech. Whisper, on the other hand, was not only the best-performing off-the-shelf 
model on English dysarthric read speech in line with~\cite{alsayegh2025zero}, it also achieved similar performance to naive human listeners for both read and spontaneous speech. 
To our knowledge, this is the first study to show that an off-the-shelf ASR system performed similarly to naive human listeners on dysarthric continuous spontaneous speech recognition, though WERs exceed 70\%.


After fine-tuning on the target speaker's speech, both personalized DSR models (FT-Whisper and FT-OmniASR) outperformed naive human listeners by a large margin, with an average absolute improvement of WER by more than 34\%, which is consistent with earlier findings where personalized DSR models outperformed experienced human listeners on Mandarin and English read dysarthric speech~\cite{sun2023cdsd, green21_interspeech}. Recent single-speaker studies on Japanese and Czech dysarthric speech similarly found that personalized DSR models outperformed naive listeners~\cite{ogasawara2026ss, nadrchal2026personalized}; however, listeners familiar with the speaker, such as family and friends or the best-performing of the speaker's assistants, still outperformed the models~\cite{ogasawara2026ss, nadrchal2026personalized}. Our results add to the increasing literature showing the potential of personalized DSR and, crucially, show its potential for supporting speech communication of people with a speech impairment, in line with recent studies integrating personalized DSR into smartphone apps~\cite{huber2026adapting, nadrchal2026personalized}. Overall, FT-Whisper achieved the best performance in this study on both read and spontaneous speech. Note that our human listeners were allowed to listen to the stimuli multiple times. Our analysis however showed that although this potentially improved the WER for the naive human listeners, the naive human listeners particularly listened to less intelligible utterances more often compared to more intelligible utterances as illustrated by the positive link between increasing listening times and higher WER.


Interestingly, the topic of the utterances played a role in recognition performance for both the naive human listeners and the ASR systems. For the naive human listeners and all off-the-shelf ASR systems, recognition performance was lower for the statistics-related read speech (customized read speech) than for the generic read speech, and in fact even lower than for the spontaneous speech. Fine-tuning on the dysarthric speech, however, removed this difference for FT-Whisper, while for FT-Omni ASR, recognition performance was even better on the customized speech than the generic read speech. These results suggest that fine-tuning models on customized read speech may improve the ability of speakers with dysarthria to communicate about their favorite topics in personalized communication support systems. Therefore, in our future work, we will continue to collect customized read speech for developing personalized DSR models for our app, to better support the day-to-day communication of speakers with dysarthria. 

Moreover, comparison of the results of the DSR systems and the naive human listeners showed that future work should focus on improving the recognition of spontaneous speech and longer read utterances as these were the speech types with the lowest performance for FT-Whisper and the naive human listeners. The phoneme error analysis showed that, for this speaker, future research should focus on improving the DSR recognition performance on a particular set of Dutch phonemes, including $/$\textipa{Z}$/$, $/$\textipa{S}$/$, $/$\textepsilon\textlengthmark$/$, $/$f$/$, $/$y$/$, $/$\textipa{N}$/$, and $/$j$/$. Although the high degree of inter-speaker variability among speakers with dysarthria suggests that these results may not generalize to other speakers with dysarthria, assessing their generalizability remains an important topic for future work. 


In conclusion, the recognition of both read and spontaneous speech of a single Dutch speaker with severe dysarthria is highly challenging for both naive human listeners and SotA off-the-shelf ASR models.
Personalized DSR models, however, substantially outperform naive human listeners. Though the performance remains far below SotA performance on typical speech, 
personalized DSR can already support the day-to-day communication of speakers with dysarthria. 

\section*{Acknowledgments}

The authors thank Wenyi Chu for her help and support with finding human listener participants.

During the preparation of this work the authors used Generative AI (ChatGPT) to improve language and readability and to debug Python and R scripts for statistical significance tests. After using this tool, the authors reviewed and edited the content as needed and take full responsibility for the content of the publication. 

\bibliographystyle{IEEEtran}
\bibliography{IEEEabrv,mybib}

\end{document}